\documentclass[letterpaper, 10 pt, conference]{ieeeconf}  
\usepackage{amsmath}
\usepackage{cite}

\usepackage{hyperref}
\usepackage{graphicx} 
\usepackage{subfig} 
\graphicspath{{./images/}} 
\usepackage{caption}
\usepackage{graphicx}
\usepackage{float} 
\usepackage{subcaption}

\IEEEoverridecommandlockouts                              

\title{\LARGE \bf
Structured Diversity Control: A Dual-Level Framework for Group-Aware Multi-Agent Coordination}

\author{Anonymous}
\author{Shuocun Yang$^{1}$, Huawen Hu$^{1}$, Xuan Liu$^{1}$,Yincheng Yao$^{1}$,Enze Shi$^{1}$, Shu Zhang$^{1, *}$ \\ 
	\thanks{*Corresponding author: Shu Zhang, shu.zhang@nwpu.edu.cn.}
	\thanks{$^{1}$ Northwestern Polytechnical University, Xi’an 710072, China.}%
}
\begin{document}

\maketitle
\thispagestyle{empty}
\pagestyle{empty}

\begin{abstract}
Controlling the behavioral diversity is a pivotal challenge in multi-agent reinforcement learning (MARL), particularly in complex collaborative scenarios. While existing methods attempt to regulate behavioral diversity by directly differentiating across all agents, they lack deep characterization and learning of multi-agent composition structures. This limitation leads to suboptimal performance or coordination failures when facing more complex or challenging tasks. To bridge this gap, we introduce Structured Diversity Control (SDC), a framework that redefines the system-wide diversity metric as a weighted combination of \textbf{intra-group diversity}, which is minimized for cohesion and \textbf{inter-group diversity}, which is maximized for specialization. The trade-off is governed by a pre-set \textbf{Diversity Structure Factor (DSF)}, allowing for fine-grained, group-aware control over the collective strategy. Our method directly constrains the policy architecture without altering reward functions. This structural definition of diversity enables SDC to deliver substantial performance gains across various experiments, including increasing average rewards by up to 47.1\% in multi-target pursuit and reducing episode lengths by 12.82\% in complex neutralization scenarios. The proposed method offers a novel analytical perspective on the problem of cooperation in group-aware multi-agent systems.
\end{abstract}

\section{INTRODUCTION}

Multi-agent reinforcement learning (MARL) has become a powerful paradigm for solving complex cooperative tasks\cite{ref11,ref12}. A central challenge in MARL is fostering effective coordination, where agents must balance individual behaviors with collective goals\cite{ref1}. While foundational techniques like parameter sharing improve sample efficiency, they often lead to policy homogenization, limiting the system's adaptability and performance in scenarios requiring diverse roles\cite{ref38,ref39}. This highlights the critical need for explicit behavioral diversity control — a mechanism to regulate not just the overall level of diversity, but its underlying structure.


\indent Behavioral diversity control has emerged as a critical research area in multi-agent coordination, garnering increasing attention in recent years. This growing interest has led to the development of various approaches for quantifying and regulating agent behavioral differences. 
Hu et al. have explored methods to measure diversity by learning latent representations of policies to compute a Multi-Agent Policy Distance (MAPD)~\cite{ref50}.
Bettini et al. introduced System Neural Diversity (SND)\cite{ref3}, a Wasserstein distance-based metric that aggregates pairwise agent distances to measure system-level behavioral heterogeneity while maintaining agent number invariance. Building upon SND, Bettini et al. further proposed DiCo~\cite{ref2}, which achieves precise diversity control by representing agent policies as combinations of shared homogeneous and agent-specific heterogeneous components with architectural constraints. Zhang et al. developed Intrinsic Action-tendency Consistency (IAM)~\cite{ref50}, integrating intrinsic rewards with Centralized Training with Decentralized Execution (CTDE) frameworks to address action-tendency disagreements through prediction-based incentive mechanisms.

Despite these advances, existing diversity control methods still face significant limitations in structured multi-agent scenarios with groupings. These pioneering methods, including both the metrics and the control frameworks built upon them, share a common limitation: their formulation is monolithic. Current approaches either promote diversity among individual agents~\cite{ref8,ref31,ref36,ref37} or enforce system-wide uniformity~\cite{ref4,ref5}, treating all agents as a single population without differentiating between desired low diversity within groups versus high diversity between groups. This monolithic nature leaves the critical challenge of controlling diversity structure unaddressed, leading to suboptimal performance in complex coordination tasks where different groups require distinct diversity characteristics—some needing tight cooperation while others requiring specialized differentiation.
\begin{figure*}[t] 
    \centering
    \includegraphics[width=\textwidth]{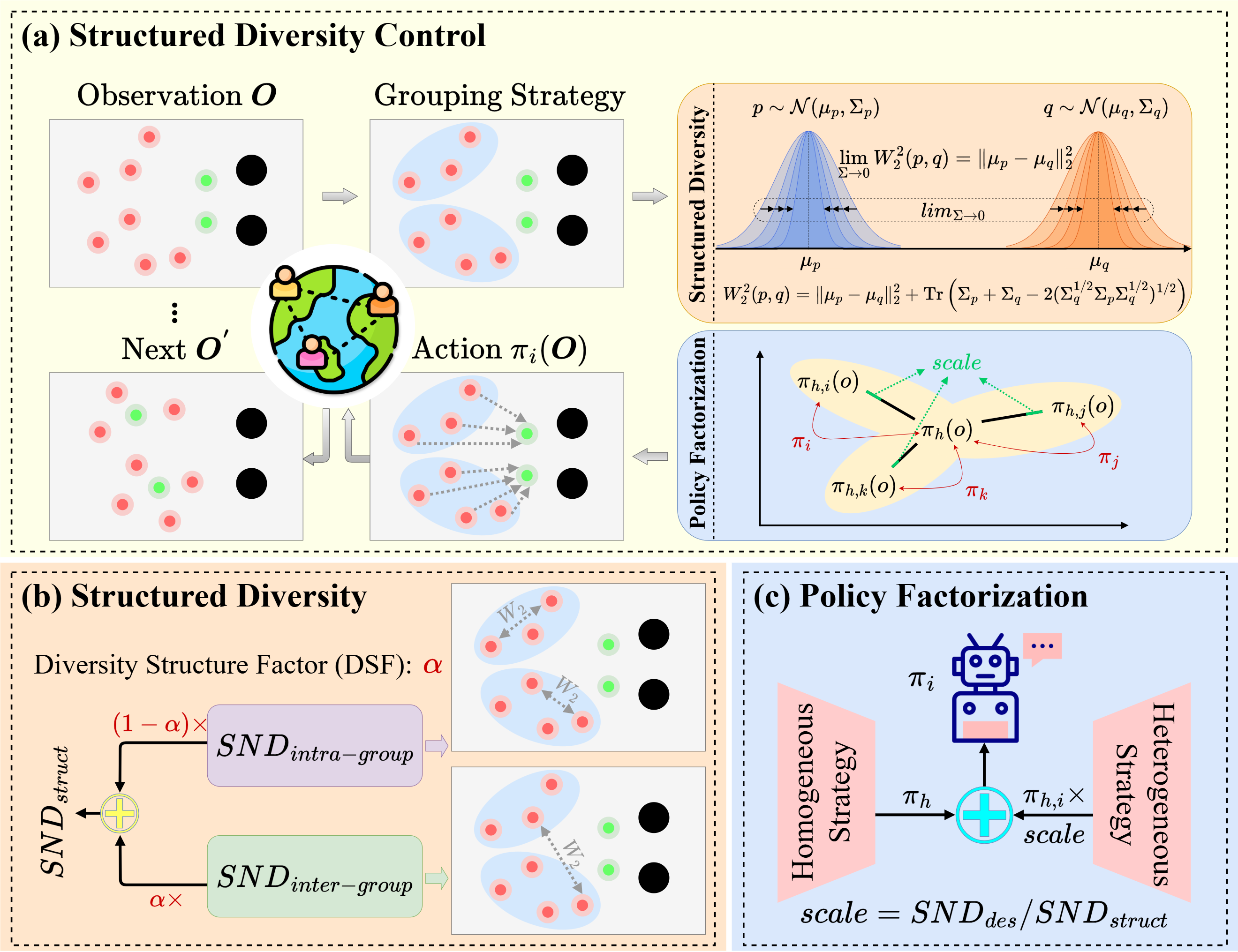} 
    \caption{Overview of the Structured Diversity Control (SDC) framework within a MARL training loop: \textbf{(a) Structured Diversity Control:} Within the SDC framework, a structured diversity constraint is imposed on the agent policies. 
\textbf{(b) Structured Diversity:} At each training step, the framework first computes the structured system diversity $\text{SND}_{\text{struct}}$ from the heterogeneous policy components. 
\textbf{(c) Policy Factorization:} Each agent's final policy is represented as the sum of a parameter-shared, homogeneous component and a rescaled, per-agent heterogeneous component. The computed $\text{SND}_{\text{struct}}$ is then compared to a desired target value $\text{SND}_{\text{des}}$ to generate a scale factor. This factor is used to rescale the heterogeneous components, resulting in the final updated policies. } 
    \label{fig:1}
\end{figure*}
To address this gap, we introduce Structured Diversity Control (SDC), a novel framework that extends the principle of architectural constraint-based control to structured multi-agent systems. SDC introduces a mechanism to explicitly manage the structure of diversity by redefining the total diversity metric. It is calculated from three key components: 

(1) \textbf{Intra-Group Diversity}, a measure of policy dissimilarity within each group, which is typically minimized to facilitate tight cooperation; 

(2) \textbf{Inter-Group Diversity}, a measure of policy dissimilarity between groups, which is encouraged to enable an effective division of labor;

(3) \textbf{Diversity Structure Factor (DSF)}, which governs the balance between these two components, allowing practitioners to inject prior knowledge about the task's coordination requirements. 

SDC offers a generalizable solution compatible with any actor-critic framework. Our core contributions are threefold:
\begin{itemize}
    \item We propose a novel dual-level framework for explicit behavioral diversity control in grouped MARL scenarios, bridging a critical gap left by monolithic control methods.
    \item We introduce the Diversity Structure Factor, a mechanism for fine-grained tuning of group-based strategies to match diverse task demands.
    \item We provide extensive results showing that by enabling more granular control over behavioral diversity, SDC achieves significantly superior task performance and more effective specialization compared to its highly competitive counterpart, DiCo, especially in complex cooperative tasks requiring structured team coordination.
\end{itemize}

\section{related work}
Behavioral diversity control in multi-agent reinforcement learning is an emerging but promising research area\cite{ref9}. Early studies have focused on single-agent scenarios, encouraging exploration by introducing entropy regularization or diversity rewards\cite{ref10,ref27}. However, it is often difficult to effectively coordinate the relationship between individuals and the whole when these approaches are directly extended to multi-agent systems\cite{ref13,ref32,ref33,ref34,ref35}.
\subsection{Diversity via Internal Mechanisms} 
Such approaches directly motivate agents to explore diverse strategies and avoid behavioral convergence by designing reward mechanisms, learning frameworks, or strategy optimization methods within the agents. By designing an inherent reward mechanism, Ben Eysenbach et al.\cite{ref16} propose a method to maximize mutual information, encourage agents to explore different strategies and avoid behavioral convergence. Through self-course learning, Igor Mordatch et al.\cite{ref20} have promoted agents to develop a variety of tool use strategies in interaction. Max Jadeberg et al.\cite{ref18} use population methods such as evolutionary algorithms or MAP-Elites. They advocate maintaining a diverse population of agents and promoting diversity through parallel training and strategy exchange.
\subsection{Diversity via External Interaction} 
This type of approach motivates agents to develop diverse strategies in their interactions with the environment or other agents by introducing external interaction mechanisms. Marcin Andrychowicz et al.\cite{ref17} introduce a competition mechanism to promote agents to develop diversified strategies in confrontation. Peng Peng et al.\cite{ref19} introduce opponent modeling modules in the Actor-Critic framework to further enhance diversity. These efforts have solved the problem of strategy convergence in MARL from different angles, provided theoretical support and practical verification for the design and optimization of complex multi-agent systems, and promoted the application of MARL in open environments, dynamic tasks, and complex interaction scenarios.\\
\indent Although existing methods have achieved significant results in controlling behavioral diversity, they primarily focus on the diversity of individual agents or the overall system\cite{ref6,ref7}, lacking a systematic analysis of behavioral diversity from a grouping perspective\cite{ref21,ref22,ref23,ref24,ref25,ref26}. Specifically, existing methods do not fully consider the need for behavioral diversity in grouped tasks and lack fine-grained control over intra-group cooperation and inter-group task allocation, making it difficult to achieve efficient task distribution and collaboration in complex scenarios\cite{ref28}. Therefore, we propose SDC, which provides a new solution for task allocation and collaboration in complex multi-agent systems by introducing two key concepts, intra-group diversity and inter-group diversity, and constraining the behavior of agents from the grouping perspective. 

\section{METHOD}
In this section, we present Structured Diversity Control (SDC), a novel framework Fig.~\ref{fig:1} designed to control behavioral diversity in multi-agent systems through explicit group-aware coordination. Our approach extends architectural constraint-based methods to address both intra-group cohesion and inter-group specialization simultaneously. We first formalize the problem setting and then elaborate on the key components of our proposed methodology.

\subsection{Problem Formulation: Grouped POMGs}
We formulate the multi-agent cooperative task as a Partially Observable Markov Game (POMG)~\cite{ref48}, defined by the tuple $\langle \mathcal{A}, \mathcal{S}, \{\mathcal{O}_i\}, \{\mathcal{A}_i\}, P, R, \gamma \rangle$. Here, $\mathcal{A}$ is the set of $N$ agents. A key characteristic of our setting is that the set of $N$ agents, $\mathcal{A}$, is partitioned into $K$ disjoint groups, denoted by $\mathcal{G} = \{G_1, G_2, \dots, G_K\}$. Each group $G_k \in \mathcal{G}$ contains $n_k$ agents, satisfying the conditions $\bigcup_{k=1}^{K} G_k = \mathcal{A}$ and $G_i \cap G_j = \emptyset$ for $i \neq j$. Each agent $i$ learns a policy $\pi_i$ that outputs a continuous action distribution, typically a multivariate Gaussian $\mathcal{N}(\mu_i, \Sigma_i)$. 

Our goal is to learn policies that not only maximize the shared reward but also adhere to a structured behavioral diversity constraint defined over these groups.
\subsection{Dual-Level Diversity Formulation}
Our core idea is to create a diversity measure that is aware of the group structure by decomposing the standard system-wide diversity metric into two semantically meaningful components. Our formulation is built upon the System Neural Diversity (SND) metric~\cite{ref3}, which provides a principled way to quantify behavioral diversity using the 2-Wasserstein distance ($W_2$) as its underlying pairwise measure. In our work, we leverage this same foundational metric but decompose it to apply in a structured, dual-level manner, capturing the nuanced coordination requirements of grouped systems.
\subsubsection{Intra-Group Diversity}
The first component of our formulation measures the cohesion within a group. For each group $G_k \in \mathcal{G}$, we define its intra-group diversity as the average pairwise SND among its members. In this regard, we follow the standard method used by DiCo~\cite{ref2} but apply it at the sub-group level. The intra-group diversity of the heterogeneous policy components, $\text{SND}_{\text{intra}}(G_k)$, is calculated as:
\begin{equation}\label{eq:intra_diversity}
    \text{SND}_{\text{intra}}(G_k) = \frac{2}{n_k(n_k-1)} \sum_{i=1}^{n_k-1}\sum_{j=i+1}^{n_k} W_2(\pi_{h,i}(o), \pi_{h,j}(o))
\end{equation}
A lower $\text{SND}_{\text{intra}}$ value reflects stronger intra-group cohesion, leading to more effective collaborative behavior.
\subsubsection{Inter-Group Diversity}
The second component explicitly measures the specialization between groups. For any two groups, $G_k$ and $G_l$, we define their inter-group diversity as the average pairwise SND between their respective members. The inter-group diversity of the heterogeneous components, $\text{SND}_{\text{inter}}(G_k, G_l)$, is:
\begin{equation}\label{eq:inter_diversity}
    \text{SND}_{\text{inter}}(G_k, G_l) = \frac{1}{n_k n_l} \sum_{i \in G_k} \sum_{j \in G_l} W_2(\pi_{h,i}(o), \pi_{h,j}(o))
\end{equation}
A higher $\text{SND}_{\text{inter}}$ value corresponds to a clearer division of labor between groups, as they adopt distinct, specialized policies.
\subsection{Structured Diversity and Policy Factorization}
SDC computes the total structured diversity, $\text{SND}_{\text{struct}}$, by structurally combining the dual-level components using the Diversity Structure Factor (DSF) $\alpha$. For a general system with $K$ groups, the total structured diversity is a weighted sum:
\begin{align}
    \text{SND}_{\text{struct}} = {} & (1-\alpha) \cdot \left( \frac{1}{K} \sum_{k=1}^{K} \text{SND}_{\text{intra}}(G_k) \right) + {} \nonumber \\
    & \alpha \cdot \left( \frac{1}{\binom{K}{2}} \sum_{k=1}^{K-1} \sum_{l=k+1}^{K} \text{SND}_{\text{inter}}(G_k, G_l) \right)
    \label{eq:total_snd}
\end{align}
where $\binom{K}{2}$ is the binomial coefficient, representing the total number of unique group pairs. A high $\alpha$ prioritizes inter-group specialization, whereas a low $\alpha$ emphasizes intra-group cohesion. This formulation is valid for any $K \ge 2$.

Inspired by prior approaches that emphasize structural constraints on policy architectures~\cite{ref2}, we represent each agent's final policy $\pi_i$ as the sum of a shared component $\pi_h$ and a rescaled per-agent deviation $\pi_{h,i}$. To enforce the desired overall diversity level, $\text{SND}_{\text{des}}$, we use a scaling factor, scale:
\begin{equation}\label{eq:policy_update}
    \pi_i(o) = \pi_h(o) + \text{scale} \cdot \pi_{h,i}(o)
\end{equation}
where the scaling factor is computed as:
\begin{equation}\label{eq:scale_factor}
    \text{scale} = \frac{\text{SND}_{\text{des}}}{\text{SND}_{\text{struct}}}
\end{equation}

This mechanism, leveraging the properties of the Wasserstein distance, guarantees that the diversity of the final policies, when measured using the same structural formulation as Eq.~\eqref{eq:total_snd}, will match the target value, $\text{SND}_{\text{des}}$, without altering the learning objective.

\section{EXPERIMENTS}

We conduct a series of experiments designed to comprehensively evaluate our proposed SDC framework. Our evaluation aims to answer three key questions: 
(1) To what extent can SDC achieve precise regulation of structured system diversity, ensuring alignment with desired targets?
(2) How does the DSF modulate the balance between intra-group cohesion and inter-group specialization in the learned policies?
(3) Does structured, group-aware diversity control yield measurable performance gains over monolithic control, particularly in tasks requiring coordinated multi-agent behaviors?

\subsection{Experimental Setup}

\subsubsection{Environments}
We use two challenging multi-agent environments designed to test structured cooperative strategies. As shown in Fig~\ref{fig:1}(a), the green and red circles denote the pursuers and escapees, respectively, while the black shapes act as static obstacles.
\paragraph{Multi-Pursuer Tag} Our first environment extends the simple tag scenario from the Multi-Agent Particle Environment (MPE). To study the impact of scale and task complexity, we establish a three-tiered difficulty system: 5 pursuers vs. 2 escapees (difficult), 6 vs. 2 (medium), and 7 vs. 2 (easy). Crucially, we implement a group collaboration mechanism where pursuers are partitioned into two teams, each tasked with capturing one of the escapees. This setup necessitates both tight intra-group cohesion  and clear inter-group specialization . The evaluation metric is the team's average reward.

\paragraph{Shielded Tag} To further test coordinated target neutralization, we introduce a variant of simple tag. In this environment, each of the $N$ escapee is assigned a shield attribute. A target is only neutralized when its shield value is reduced to zero by the pursuers. The episode terminates when all targets are neutralized. The primary evaluation metric is the episode length; a shorter episode indicates a more efficient strategy. This task explicitly rewards "focus fire" where high intra-group cohesion and high inter-group specialization are paramount for success.

\subsubsection{Implementation Details}
In our experiments, we partition the pursuer agents into distinct groups, with the number of groups set to match the number of escapees. All agents are trained using the IPPO algorithm~\cite{ref37}. For experiments, we train for at least 12 million environment steps across 600 parallel environments, using at least 4 distinct random seeds to ensure reliability and reproducibility. Our primary baseline for comparison is DiCo~\cite{ref2}, the highly competitive method for architectural diversity control.

\subsection{Verification of Diversity Control Capability}
\begin{figure}[h] 
    \centering
    \includegraphics[width=1.0\columnwidth]{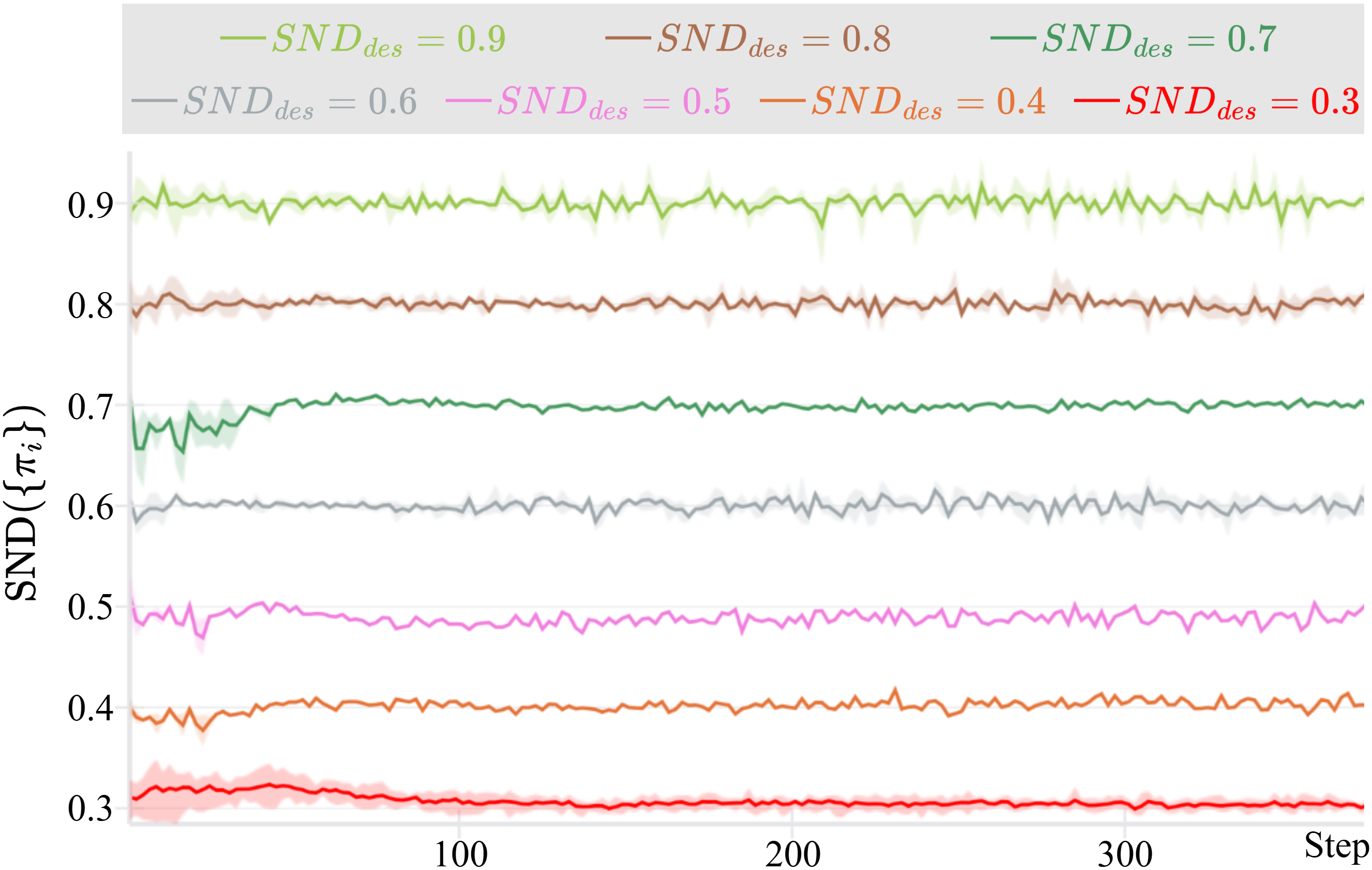} 
    \caption{Verification of SDC's diversity control capability.  Each colored line represents a separate training run with a different target diversity level, $\text{SND}_{\text{des}}$, ranging from 0.3 to 0.9, at a fixed DSF of $\alpha=0.9$.}
    \label{fig:diversity_control_verification}
\end{figure}

First, we verify that SDC can accurately control the total system diversity $\text{SND}(\{\pi_i\})$ to a desired level, $\text{SND}_{\text{des}}$. 
To this end, we fix the DSF to a representative value of $\alpha=0.9$ and train a series of models with varying target diversity levels, $\text{SND}_{\text{des}} \in \{0.3, 0.4, \dots, 0.9\}$. 
The results are presented in Fig.~\ref{fig:diversity_control_verification}.

As the Fig.~\ref{fig:diversity_control_verification} clearly illustrates, for each specified target value, the measured system diversity of the final policies rapidly converges to and then stably oscillates around its target level throughout the training process. 
For instance, the run with $\text{SND}_{\text{des}}=0.3$ (red line) quickly settles near the 0.3 level, while the run with $\text{SND}_{\text{des}}=0.9$ (light green line) consistently maintains its diversity around 0.9. 
This provides strong evidence that our structural redefinition of the SND metric does not compromise the fundamental control capability of the architectural constraint approach, validating SDC as a reliable method for precise diversity control.
\begin{figure*}[t] 
    \centering
    \includegraphics[width=1\textwidth]{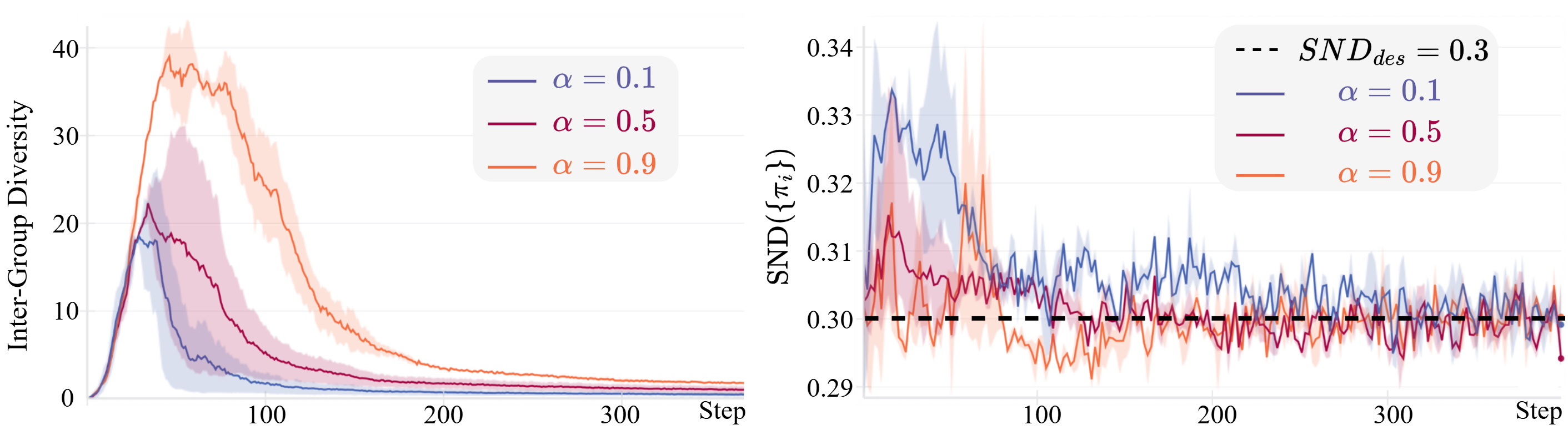} 
    \caption{Analysis of the Diversity Structure Factor (DSF, $\alpha$). (\textbf{Left}) The effect of $\alpha$ on the inter-group diversity of the heterogeneous components. Higher $\alpha$ values, which place more weight on inter-group diversity, lead to a correspondingly higher and more sustained level of specialization between groups. (\textbf{Right}) Verification of total diversity control under different DSF settings. For a fixed target diversity of $\text{SND}_{\text{des}}=0.3$, the measured system diversity $\text{SND}(\{\pi_i\})$ remains stable and accurate across all tested $\alpha$ values, demonstrating the robustness of our control mechanism.
    } 
    \label{fig:dsf_analysis}
\end{figure*}
\subsection{Analysis of the Diversity Structure Factor (DSF)}
In this section, we analyze the effect of the DSF($\alpha$), which is the core of our structural control mechanism. We conduct ablation studies with representative DSF values of $\alpha \in \{0.1, 0.5, 0.9\}$. Across these settings, the system maintains robust control over the total diversity. 
\begin{figure*}[t] 
    \centering
    \includegraphics[width=1\textwidth]{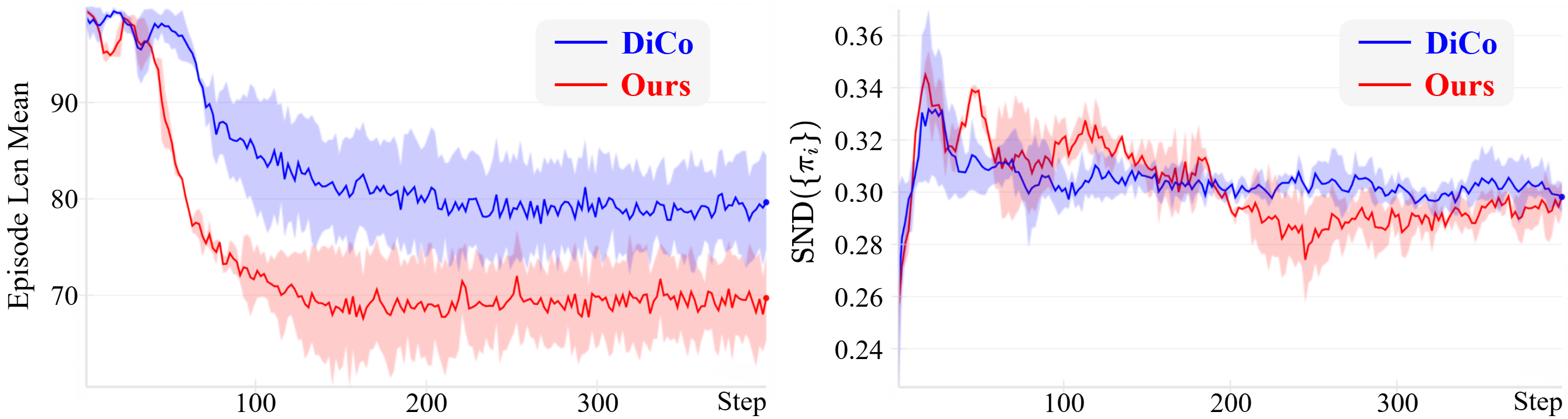} 
    \caption{Performance and diversity comparison in the Shielded Tag environment (7vs2 scenario) with a target diversity of $\text{SND}_{\text{des}}=0.3$. (\textbf{Left}) Mean episode length over training. SDC (red) learns a superior strategy, completing the task significantly faster than DiCo (blue). (\textbf{Right}) Measured total system diversity, $\text{SND}(\{\pi_i\})$. Both methods successfully converge to and maintain the target diversity level of 0.3.}
\label{fig:shielded_tag_results} 
    \label{fig:9}
\end{figure*}
The results, presented in Fig.~\ref{fig:dsf_analysis}, confirm that the DSF acts as an intuitive and effective control knob for tuning the collective strategy. The left panel of Fig.~\ref{fig:dsf_analysis} shows the inter-group diversity of the heterogeneous components throughout training. As hypothesized, a higher value of $\alpha$ directly correlates with a higher and more sustained level of inter-group diversity. This demonstrates that by increasing the weight on the inter-group component in our structured diversity metric, SDC successfully encourages greater specialization and a more pronounced division of labor between the agent groups.

Crucially, this fine-grained structural control does not compromise the overall stability of the system. The right panel of Fig.~\ref{fig:dsf_analysis} shows the measured total system diversity for a fixed target of $\text{SND}_{\text{des}}=0.3$ across the different $\alpha$ settings. The plot clearly indicates that SDC maintains robust and accurate control over the total diversity, regardless of how the internal diversity structure is weighted. This validates that the DSF provides a direct and interpretable mechanism for practitioners to inject task-specific priors about the desired balance between cohesion and specialization, without sacrificing the precision of the overall diversity control.
\subsection{Performance Comparison in Group-Based Tasks}
Having validated SDC's control mechanisms, we now evaluate its performance in tasks that benefit from structured coordination.

\begin{figure*}[t] 
    \centering
    \includegraphics[width=1\textwidth]{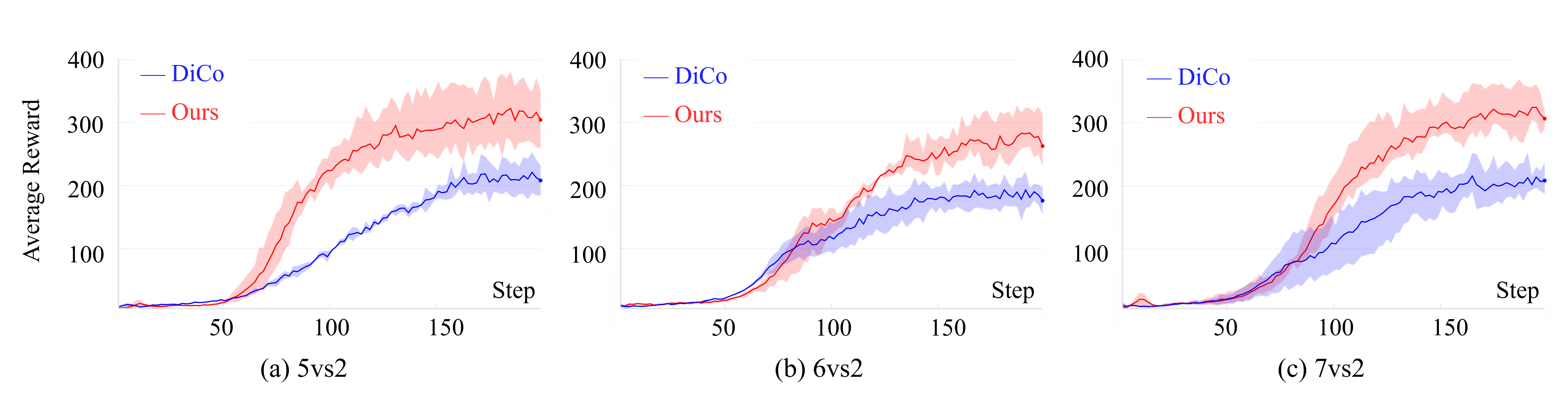} 
    \caption{Comparison of average rewards between DiCo and SDC across scenarios with a target diversity of $\text{SND}_{\text{des}}=0.3$. (a) Comparison of Average Rewards for DiCo and SDC in “5 chasing 2" scenario. (b) Comparison of Average Rewards for DiCo and SDC in “6 chasing 2" scenario. (c) Comparison of Average Rewards for DiCo and SDC in “7 chasing 2" scenario.
    } 
    \label{fig:2}
\end{figure*}
\subsubsection{Results in Shielded Tag}

In the Shielded Tag environment, where the need for structured coordination is even more critical, SDC's advantage becomes more pronounced. For this experiment, we set a target diversity of $\text{SND}_{\text{des}}=0.3$ for both methods. As illustrated in the left panel of Fig.~\ref{fig:shielded_tag_results}, SDC agents learn significantly more efficient strategies, achieving consistently shorter episode lengths compared to DiCo agents. This indicates a faster neutralization of all targets.
The strategic superiority of SDC stems from its ability to foster an optimal "focus fire" strategy, which this task explicitly rewards. SDC's structured control mechanism, governed by the DSF, can be configured to simultaneously encourage high intra-group cohesion  and high inter-group specialization. In contrast, DiCo's monolithic constraint struggles to find this delicate balance. It often results in agents spreading their attacks too thinly across multiple targets, which is a highly inefficient way to overcome the shield mechanic, thus leading to longer episodes. 

Crucially, the right panel of Fig.~\ref{fig:shielded_tag_results} provides a critical piece of evidence: the observed performance gap is not due to a failure in diversity control. Both SDC and DiCo successfully steer their total system diversity to the specified target of 0.3 and maintain it throughout training. This isolates the source of the performance difference, providing compelling evidence that SDC's superiority is a direct result of its ability to organize diversity into a more effective, group-aware structure, which is essential for solving complex, structured cooperative tasks.

\begin{figure}[htbp]
	\centering
	\includegraphics[width=1\linewidth]{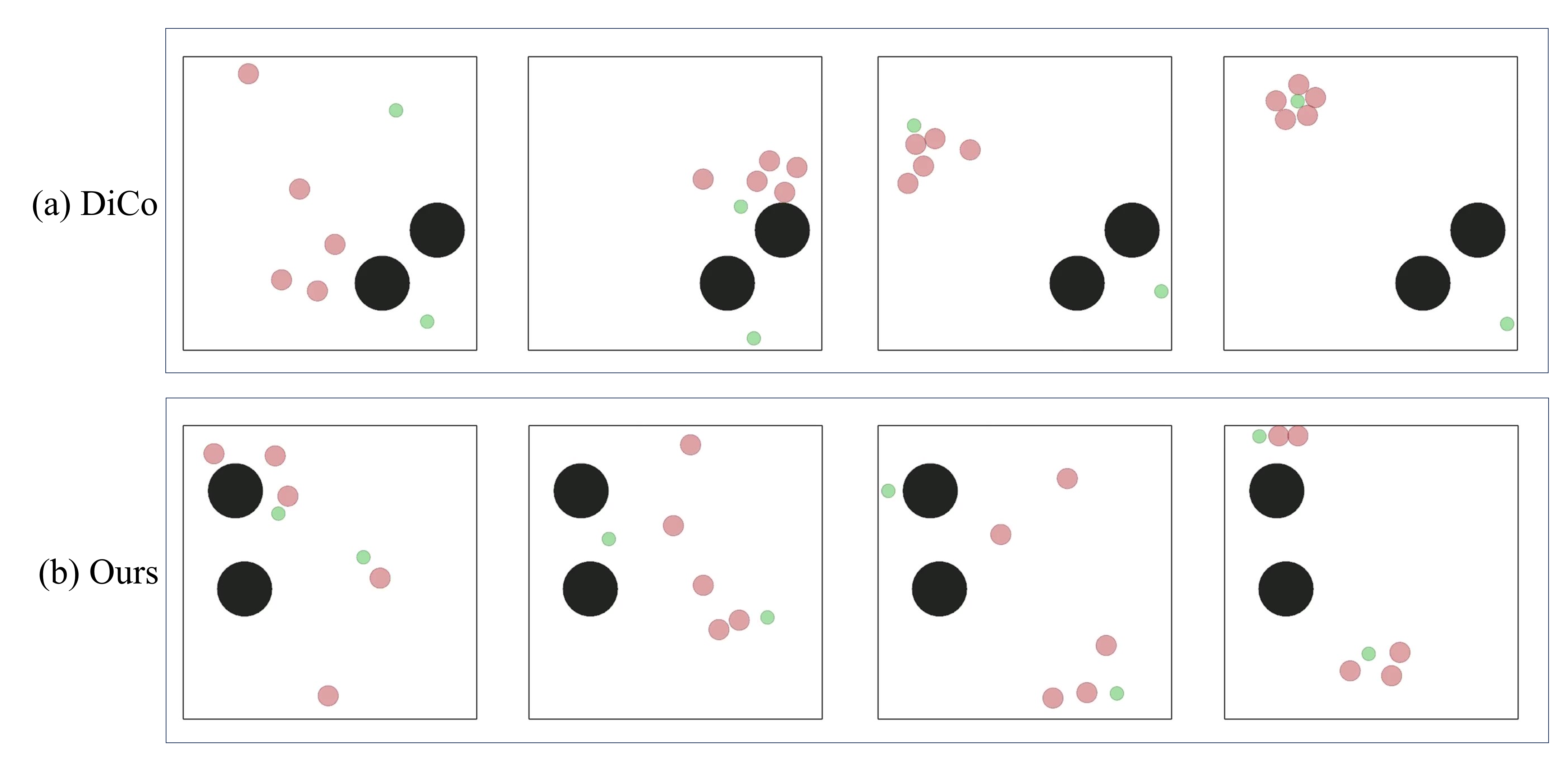}
	\caption{DiCo and SDC visualization process in “5 chasing 2" scenario. The four frames correspond to 0\%, 25\%, 75\%, and 100\% of the episode's total duration, respectively.}
	\label{fig:4}
\end{figure}
\begin{figure}[htbp]
	\centering
	\includegraphics[width=1\linewidth]{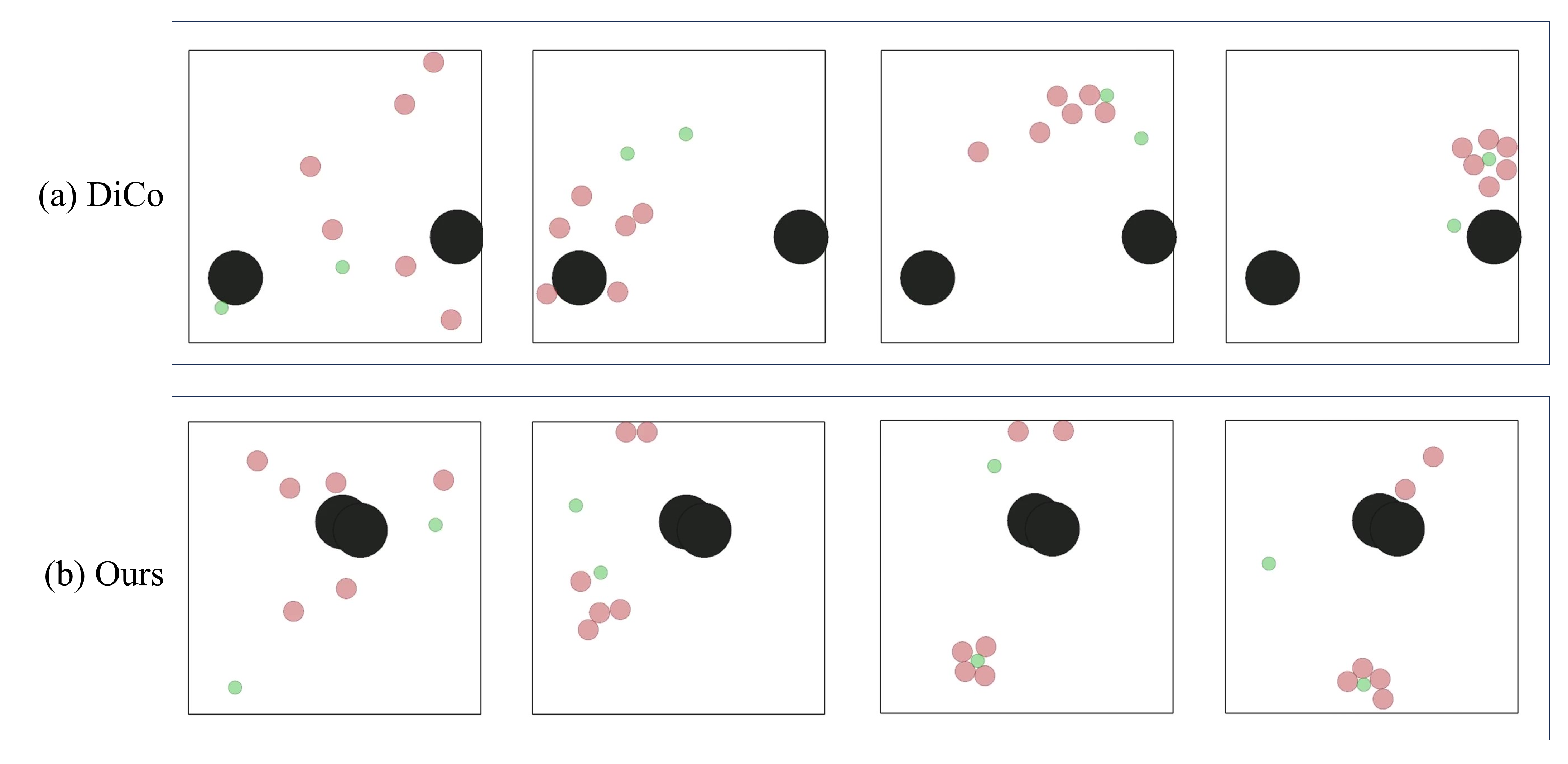}
	\caption{ DiCo and SDC visualization process in “6 chasing 2" scenario. The four frames correspond to 0\%, 25\%, 75\%, and 100\% of the episode's total duration, respectively.}
	\label{fig:5}
\end{figure}
\begin{figure}[htbp]
	\centering
	\includegraphics[width=1\linewidth]{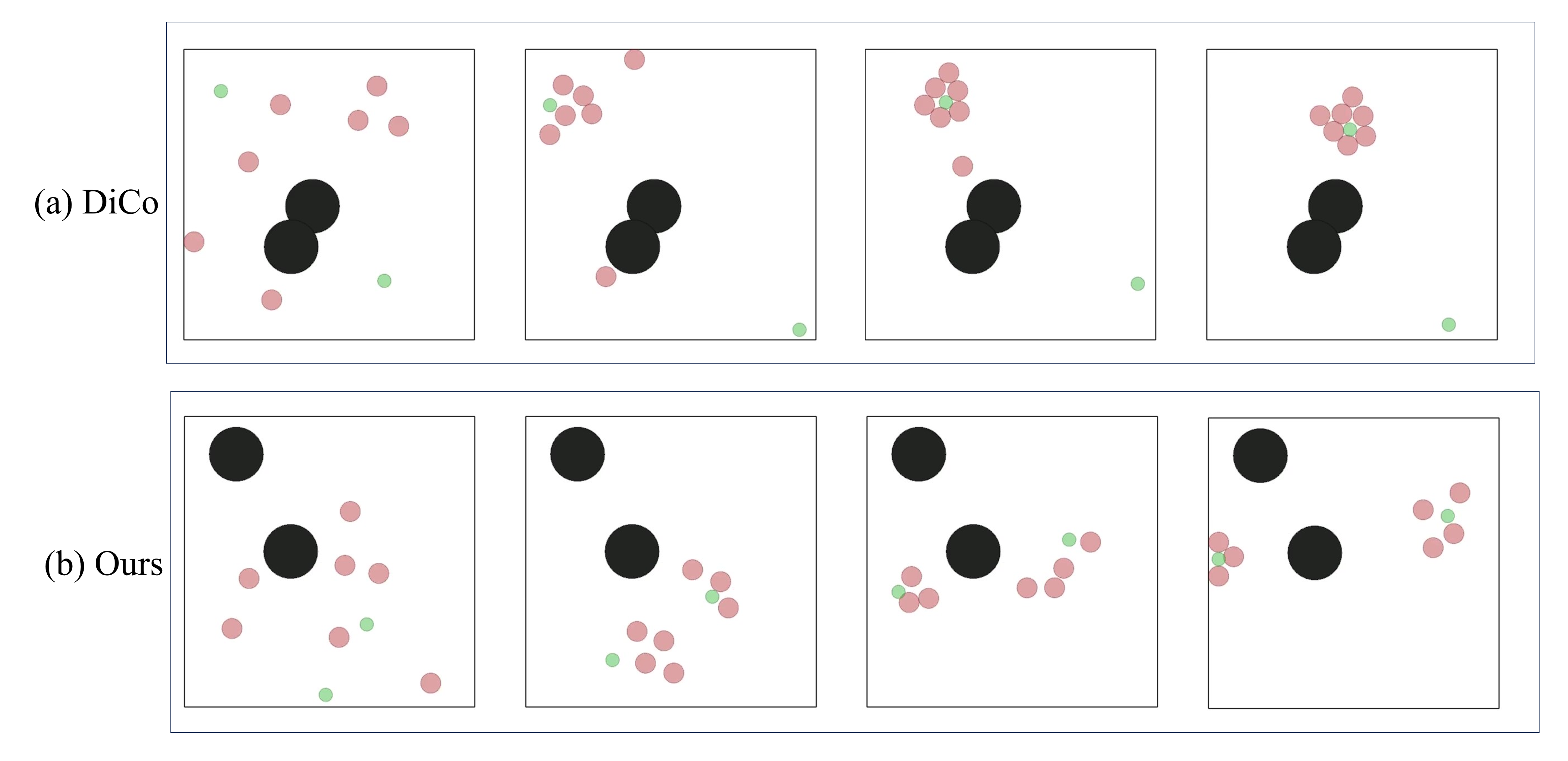}
	\caption{ DiCo and SDC visualization process in “7 chasing 2" scenario. The four frames correspond to 0\%, 25\%, 75\%, and 100\% of the episode's total duration, respectively.}
	\label{fig:6}
\end{figure}
\begin{figure}[htbp]
   \centering
    \includegraphics[width=0.5\textwidth]{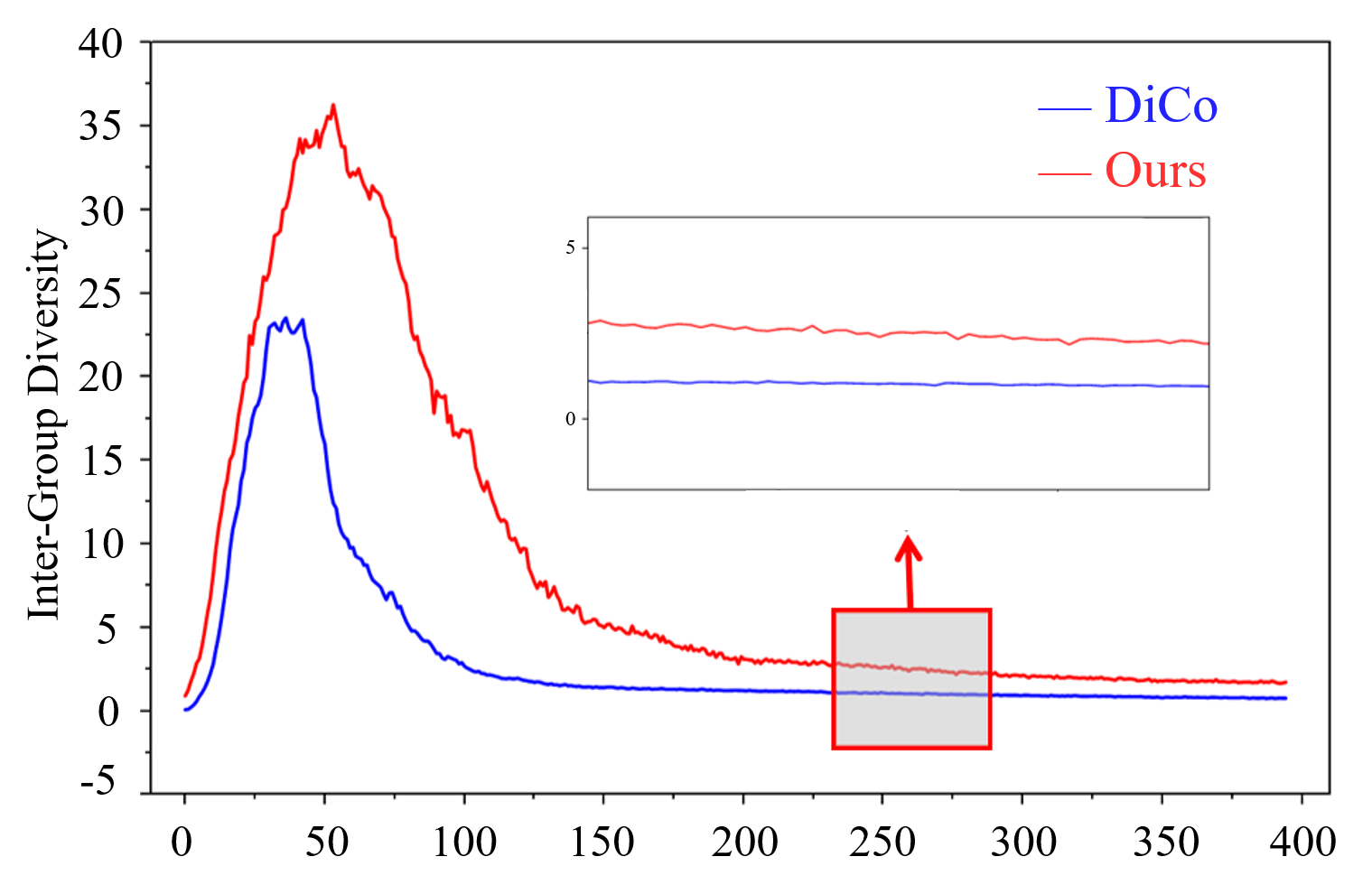} 
    \caption{Comparison of inter-group diversity between DiCo and SDC with the same target diversity.} 
    \label{fig:3}
\end{figure}
\begin{figure}[htbp]
   \centering
    \includegraphics[width=0.6\textwidth]{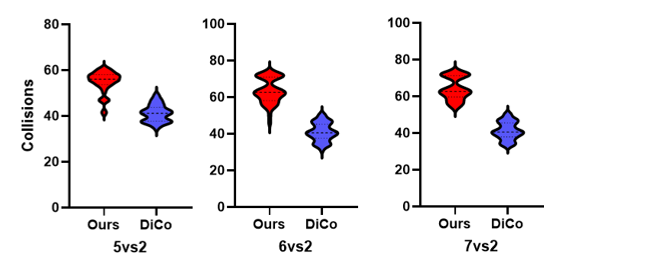} 
    \caption{Quantitative comparison of successful captures in the Multi-Pursuer Tag scenarios.}
\label{fig:collision_analysis} 
\end{figure}
\subsubsection{Results in Multi-Pursuer Tag}
In the Multi-Pursuer Tag environment, as shown by the reward curves in Fig.~\ref{fig:2}, SDC consistently achieves higher average rewards across all three difficulty settings (5vs2, 6vs2, and 7vs2). SDC demonstrates a significant performance advantage over the monolithic control of DiCo.

The underlying reason for this superiority is revealed through qualitative analysis of the learned behaviors. 
The trajectory visualizations in Fig.~\ref{fig:4} (5vs2), Fig.~\ref{fig:5} (6vs2), and Fig.~\ref{fig:6} (7vs2) consistently show a stark contrast in strategy. 
DiCo's agents, governed by a single global diversity constraint, frequently exhibit a "collapsing" behavior, where all pursuers myopically chase a single escapee. This leaves the second target completely uncontested, resulting in a highly suboptimal team strategy.

In stark contrast, SDC agents successfully learn to execute a structured, two-group strategy. 
As seen in the visualizations, they effectively partition themselves to simultaneously pursue both targets. 
This emergent division of labor is a direct result of our structured diversity control. 
To further validate this, we plot the inter-group diversity of the heterogeneous components in Fig.~\ref{fig:3}. 
The results confirm that SDC maintains a significantly higher level of inter-group diversity compared to DiCo. 
By enabling this structured control that encourages inter-group specialization, SDC facilitates a more effective division of labor, which is critical for success in this multi-target scenario. 
This strategic advantage becomes even more pronounced as the number of pursuers increases (Fig.~\ref{fig:6}), where SDC's ability to organize agents into cohesive pursuit groups prevents redundant chasing and maximizes spatial coverage, leading to better performance.

To provide a quantitative basis for this strategic difference, we further analyze the number of collisions per episode, which in this context represent successful captures of the escapees. As illustrated in Fig.~\ref{fig:collision_analysis}, the violin plots show the distribution of total collisions per episode for SDC (red) and DiCo (blue). SDC achieves a significantly higher number of collisions than DiCo across all scenarios, increasing the average collision count from approximately 41.7 to 61.3, which corresponds to a 47.0\% improvement. This discrepancy stems directly from the observed strategies: SDC's effective division of labor allows its two groups to engage both targets in parallel, creating more opportunities for captures within the same timeframe, directly explaining the higher average rewards reported in Fig.~\ref{fig:2}.

\section{CONCLUSION}
In this paper, we introduced Structured Diversity Control (SDC), a novel framework for structured behavioral diversity control in grouped multi-agent reinforcement learning that simultaneously achieves cohesion and specialization. SDC decomposes system-wide diversity into intra-group and inter-group components through architectural constraints controlled by the Diversity Structure Factor, enabling fine-grained control over group coordination without modifying learning objectives. 

Extensive experimental evaluation across multiple cooperative scenarios demonstrates that SDC achieves superior performance compared to strong baselines while maintaining precise diversity control and interpretable group structures. These results validate the importance of structured diversity control for complex multi-agent coordination tasks. SDC holds significant promise for applications in scenarios requiring group-level team coordination, including multi-robot logistics, autonomous drone swarms, intelligent manufacturing, and distributed sensing systems. The structured diversity control principles validated in this work provide a promising pathway toward solving large-scale, heterogeneous coordination problems, potentially enabling breakthrough applications across diverse domains and facilitating the transition of multi-agent systems from theoretical foundations to practical engineering deployment.

For future work, we will focus on exploring dynamic DSF adaptation mechanisms, designing data-driven grouping strategies, and validating the proposed framework in real-world scenarios such as multi-robot coordination and UAV swarms. We believe that the structural diversity control principle established in this work provides a solid foundation for advancing more sophisticated and effective multi-agent strategies.

\section{ACKNOWLEDGMENT}

This work was supported by National Natural Science Foun dation of China (62376219); Faculty Construction Project (Grant No. 25SH02010044).

\bibliographystyle{ieeetr}
\bibliography{IEEEtranBST/IEEEexample}

\end{document}